\pdfoutput=1
\documentclass{article}

\usepackage{microtype}
\usepackage{graphicx}
\usepackage{subcaption}
\usepackage{booktabs} %

\usepackage{hyperref}

\usepackage[group-separator={,}]{siunitx}

\usepackage[accepted]{icml2025}

\usepackage{amsmath}
\usepackage{amssymb}
\usepackage{mathtools}
\usepackage{amsthm}

\usepackage{tabularx}
\usepackage{adjustbox}
\usepackage{makecell}

\usepackage[capitalize,noabbrev]{cleveref}

\usepackage{xspace}
\usepackage{bm} %
\usepackage{nicefrac}       %
\usepackage{natbib}

\usepackage{url}
 
\usepackage{./stylefiles/ml-defs}

\newcommand{\xlstmlarge}{{xLSTM\,7B}\xspace}  %
\newcommand{\cellstate}[1]{#1}  %
\newcommand{\expGate}[1]{#1}  %
\newcommand{\gates}[1]{#1}  %

\theoremstyle{plain}

\theoremstyle{definition}

\theoremstyle{remark}

\usepackage{todonotes}
\usepackage{marginnote}
\makeatletter
\renewcommand{\@todonotes@drawMarginNoteWithLine}{%
\begin{tikzpicture}[remember picture, overlay, baseline=-0.75ex]%
    \node [coordinate] (inText) {};%
\end{tikzpicture}%
\marginnote[{%
    \@todonotes@drawMarginNote%
    \@todonotes@drawLineToLeftMargin%
}]{%
    \@todonotes@drawMarginNote%
    \@todonotes@drawLineToRightMargin%
}%
}
\makeatother

\icmltitlerunning{xLSTM-7B}

\begin{document}

\twocolumn[

\icmltitle{xLSTM 7B: A Recurrent LLM for Fast and Efficient Inference}

\icmlsetsymbol{equal}{*}

\begin{icmlauthorlist}
\icmlauthor{Maximilian Beck}{equal,nxai,jku}
\icmlauthor{Korbinian P\"oppel}{equal,nxai,jku}
\icmlauthor{Phillip Lippe}{equal,nxai,google}
\icmlauthor{Richard Kurle}{nxai}
\icmlauthor{Patrick M. Blies}{nxai}
\icmlauthor{G\"unter Klambauer}{nxai,jku}
\icmlauthor{Sebastian B\"ock}{nxai}
\icmlauthor{Sepp Hochreiter}{nxai,jku}
\end{icmlauthorlist}

\icmlaffiliation{nxai}{NXAI GmbH, Linz, Austria}
\icmlaffiliation{jku}{JKU, Johannes Kepler University, Linz, Austria}
\icmlaffiliation{google}{Now at Google Deepmind}

\icmlcorrespondingauthor{Maximilian Beck}{\href{mailto:maximilian.beck@nx-ai.com}{maximilian.beck@nx-ai.com}}
\icmlcorrespondingauthor{Korbinian P\"oppel}{\href{mailto:korbinian.poeppel@nx-ai.com}{korbinian.poeppel@nx-ai.com}}
\icmlcorrespondingauthor{Sebastian B\"ock}{\href{mailto:sebastian.boeck@nx-ai.com}{sebastian.boeck@nx-ai.com}}

\icmlkeywords{Machine Learning, ICML}

\hyphenation{mLSTM xLSTM sLSTM LSTM}

\vskip 0.3in
]

\printAffiliationsAndNotice{\icmlEqualContribution} %

\begin{abstract}

Recent breakthroughs in solving reasoning, math and coding problems with Large Language Models (LLMs) have been enabled by investing substantial computation budgets at inference time.
Therefore, inference speed is one of the most critical properties of LLM architectures, and there is a growing need for LLMs that are efficient and fast at inference. 
Recently, LLMs built on the xLSTM architecture have emerged as a powerful alternative to Transformers, offering linear compute scaling with sequence length and constant memory usage, both highly desirable properties for efficient inference.
However, such xLSTM-based LLMs have yet to be scaled to larger models and assessed and compared with respect to inference speed and efficiency.
In this work, we introduce \xlstmlarge, a 7-billion-parameter LLM that combines xLSTM’s architectural benefits 
with targeted optimizations for fast and efficient inference.
Our experiments demonstrate that \xlstmlarge achieves performance on downstream tasks comparable to other similar-sized LLMs, while providing significantly faster inference speeds and greater efficiency compared to Llama- and Mamba-based LLMs.
These results establish \xlstmlarge as the fastest and most efficient 7B LLM, offering a solution for tasks that require large amounts of test-time computation.
Our work highlights xLSTM's potential as a foundational architecture for methods building on heavy use of LLM inference. Our model weights, model code and training code are open-source. \\
{\fontsize{8pt}{7pt}\selectfont Model: \hspace{-0.2cm} {\fontsize{7.5pt}{7pt}\selectfont \url{https://huggingface.co/NX-AI/xLSTM-7b}} Code: \url{https://github.com/NX-AI/xlstm} and \url{https://github.com/NX-AI/xlstm-jax}}.

\end{abstract}

\newpage

\todototoc

\section{Introduction}\label{sec:introduction}

Recent breakthroughs in test-time compute scaling have unlocked significant improvements in solving complex reasoning and math problems. 
By sampling multiple promising solutions, the best answers can be provided to the user or used as training targets \cite{Yao2023TreeOfThoughts, Hao2023ReasoningLM, Guan2025rStarMath}. 
However, as state-of-the-art models such
as OpenAI o1\footnote{\tiny\url{https://openai.com/index/introducing-openai-o1-preview/}} and DeepSeek-R1 \citep{DeepseekAI2025R1} 
leverage these methods to push the capabilities of language models to new heights, the significantly increased computational overhead of test-time compute methods requires more efficient architectures that provide greater inference speeds. 
A promising path involves linear recurrent neural networks with gating mechanisms, including GLA \cite{Yang2024GLA}, Mamba \cite{Gu2024Mamba, Dao2024Mamba2}, RWKV \cite{Peng2023RWKV4, Peng2024RWKV6}, RetNet \cite{Sun2023RetNet}, and xLSTM \cite{Beck2024xLSTM}. 
Compared to Transformers, these models offer a parallel mode for efficient training \citep[e.g.][]{Yang2024GLA} and a recurrent generation mode that both scale linearly with context length.
The increased compute efficiency combined with constant memory usage during inference allows spending more compute at test-time, but also enables running models locally on edge devices acting as an interface to the user with fast response times.

xLSTM has shown competitive performance compared to alternative recurrent models and even Transformers in a controlled experimental setting using the same data and similar parameter counts \cite{Beck2024xLSTM}. 
Moreover, this architecture also excelled in other domains, such as computer vision \cite{Alkin2024VisionxLSTM}, robotics \cite{Schmied2024largerecurrentactionmodel}, molecular biology \cite{Schmidinger2024BioxLSTM}, and time series \cite{Kraus2024xLSTM-Mixer}.
However, so far, xLSTM has not been scaled to datasets beyond 300B tokens and 1.3B parameters. 
It therefore remains uncertain whether this architecture can match the Transformer's ability to scale effectively with larger model sizes and extract meaningful patterns from ever-larger datasets.

In this work, we scale the xLSTM to 7B parameters and present our \xlstmlarge, a large language model trained on 2.3T tokens from the DCLM dataset \cite{Li2024DCLM} with context length 8192 using 128 H100 GPUs. 
To achieve this, we improve and optimize the initial xLSTM architecture from \citet{Beck2024xLSTM} for optimal training efficiency and stability, 
without sacrificing performance in downstream tasks. 
Our new architecture fully relies on mLSTM cells with parallel training mode to achieve maximum speed at high language modeling performance.
We further optimize the throughput by modifying the surrounding block architecture. 
By operating the mLSTM in a lower dimensional space and adding position-wise feedforward MLP layers similar to the default Transformer blocks, we increase the amount of compute spent for highly optimized linear layers. 
Additionally, we discard components such as channel-wise convolutions or learnable skip connections to increase the GPU utilization during training.
We find that this optimized block architecture has a $2\times$ to $4\times$ higher token throughput compared to the previous xLSTM architecture of \citet{Beck2024xLSTM}, while achieving similar performance on language modeling.
In addition to the efficiency optimizations, we optimize the new xLSTM architecture for improved training stability, focusing specifically on the gating mechanism of the mLSTM cell. 
By introducing soft-capping for input and forget gates and improved initializations for the input gate we effectively mitigate high gradient norm spikes and variance, and improve the performance of our \xlstmlarge. 

In our evaluations on language downstream and long-context tasks, \xlstmlarge shows comparable performance to Transformers and Mamba models of the same size,
but with our optimized block architecture it achieves the highest prefill and generation throughput with the lowest GPU memory footprint on our inference efficiency benchmarks. 

To summarize, in this work we present targeted modifications to the xLSTM architecture in order to (i) improve training and inference efficiency, and (ii) ensure training stability at large scales.
(iii) We introduce a new language model with 7B parameters based on the xLSTM architecture trained on 2.3 T tokens with 8k context length demonstrating the highest inference speed and efficiency in our benchmarks.

We release the pre-trained model \xlstmlarge on Huggingface\footnote{ \url{https://huggingface.co/NX-AI/xLSTM-7b}} and provide the model implementation and training code \footnote{\url{https://github.com/NX-AI/xlstm-jax}} including optimized triton kernels \footnote{\url{https://github.com/NX-AI/mlstm_kernels}} for fast training and inference.

\section{Background: xLSTM with Matrix Memory}
\label{sec:background}
In this section, we reassess the mLSTM~\citep{Beck2024xLSTM}, on which we build our \xlstmlarge. 
The mLSTM cell is fully parallelizable, and, therefore, enables highly efficient large-scale model training while maintaining fast recurrent inference with constant memory.

\paragraph{Generation Mode.} 
During inference, when generating tokens, the mLSTM cell processes the series of input vectors $\Bx_t \in \dR^{d}$ for time steps $t \in \{1, \dots, T \}$ in a recurrent manner, mapping a state $\left(\Bh_{t-1}, \BC_{t-1}, \Bn_{t-1}, m_{t-1}\right)$ to a successor state $\left(\Bh_{t}, \BC_{t}, \Bn_{t}, m_{t}\right)$ given an input $\Bx_t$. 
Here, $\Bh_{t} \in \dR^{d_{hv}}$ denotes the hidden state, $\BC_{t} \in \dR^{d_{qk} \times d_{hv}}$ denotes the cell state responsible for long-term memory, $\Bn_{t} \in \dR^{d_{qk}}$ denotes the normalizer state, and $m_{t} \in \dR$ denotes the max state controlling the magnitude of the exponential input gate. 

In the recurrent mode (generation), the mLSTM cell 
\begin{equation}
    \label{eq:mlstm_rec}
    \Bh_{t} = \mathrm{mLSTMCell}\left(\Bx_{t}, \Bh_{t-1}, \BC_{t-1}, \Bn_{t-1}, m_{t-1}\right),
\end{equation}
is defined by the following state update equations:
\begin{align}
    \cellstate{m_t}                          & = \max\left\{ \gates{\log \sigma(\tilde{\Rf}_t)} + \cellstate{m_{t-1}}, \ \tilde{\Ri}_t \right\},           & \label{eq:mlstm_rec_begin}\\
    \cellstate{\BC_t} \                      & = \  \gates{\Rf_t} \ \cellstate{\BC_{t-1}} \ + \
    \expGate{\ \Ri_t} \ {\Bk_t \ \Bv_t^\top}, &                                                                                                              \\
    \cellstate{\Bn_t \ \!} \                 & = \  \gates{\Rf_t} \ \cellstate{\Bn_{t-1} \ \! } \ + \
    \expGate{\ \Ri_t} \ {\Bk_t}, \\
    \widetilde{\Bh}_t &= \frac{\cellstate{\BC_t^\top} \ \left({\Bq_t} / \sqrt{d_{qk}} \right)} { \ \max \left\{ \ABS{\cellstate{\Bn_t^\top} \ \left({\Bq_t} / \sqrt{d_{qk}} \right) }, \exp(-m_t) \right\}}, \label{eq:mlstm_rec_htilde}\\
    \Bh_t  \                                 & = \ \gates{\bfo_t} \ \odot  \ \mathrm{Norm} ( \, \widetilde{\Bh}_t \, ).   \label{eq:mlstm_rec_hidden_state_output}                                                
\end{align}
The gate activations are computed as:
\begin{align}
    \gates{\Rf_t}                            & = {\exp \left( \log \sigma(\tilde{\Rf}_t) + m_{t-1} - m_t \right) \, },                                     & \\
    \expGate{\ \Ri_t}                        & = {\exp(\tilde{\Ri}_t - m_t)},                                                                                \\
    \gates{\bfo_t} \                         & = \ \sigma
    \left( \tilde{\bfo}_t \right). \label{eq:mlstm_rec_end}
\end{align}
The query, key, and value vectors $\Bq_t,\Bk_t \in \dR^{d_{qk}}, \, \Bv_t \in \dR^{d_{hv}}$ are computed as $\{\Bq_t,\Bk_t,\Bv_t\} = \ \BW_{\{q,k,v\}} \ \Bx_t \ + \ \Bb_{\{q,k,v\}}$. 
The scalar input and forget gates $\Ri_t, \Rf_t \in \dR$ are computed from the pre-activations $\{\tilde{\Ri}_t, \tilde{\Rf}_t\} = \ \Bw^\top_{\{\Ri, \Rf\}} \ \Bx_t \ + \ b_{\{\Ri, \Rf\}}$ and the vector output gate $\bfo_t \in \dR^{d_{hv}}$ is computed from the pre-activation $\tilde{\bfo}_t  \ = \ \BW_{\bfo} \ \Bx_t \ + \ \Bb_{\bfo}$ with the sigmoid function $\sigma$.
The normalization layer $\mathrm{Norm}$ in (\ref{eq:mlstm_rec_hidden_state_output}) can be either RMSNorm~\cite{Zhang2019Rmsnorm} or LayerNorm~\cite{Ba2016LayerNorm}.

\paragraph{Training Mode.}
In training, the mLSTM cell processes a full sequence of input vectors $\BX \in \dR^{T \times d}$ and computes the hidden states $\BH \in \dR^{T \times d_{hv}}$ for all time steps $T$ in parallel. 
We denote the mLSTM cell in parallel mode (training) as
\begin{equation}
    \BH = \mathrm{mLSTMCell}\left(\BX \right).
\end{equation}
Due to the linear nature of the recurrence in equations~\eqref{eq:mlstm_rec_begin}-\eqref{eq:mlstm_rec_end}, the hidden states $\BH$ can be computed in chunks without materializing the intermediate memory states $\left(\BC_{t}, \Bn_{t}, m_{t}\right)$.
This \emph{chunkwise-parallel} form enables highly efficient training kernels, analogous to FlashLinearAttention \cite{Yang2024GLA, Yang2024FLA}, surpassing the training speeds of FlashAttention~\citep{dao:23flashattention2, shah:24flashattention3}. 
For details on the chunkwise-parallel training kernels for the mLSTM cell, we refer to~\citet{Anonymous2025TiledFlashLinearAttention}.

\paragraph{Multi-Head mLSTM.}
Similar to multi-head attention in Transformers~\citep{Vaswani2017Attention}, the xLSTM has $N_\text{head} = d / d_{hv}$ different mLSTM cells $\mathrm{mLSTMCell}^{(i)}$. 
The hidden states $\BH^{(i)}$ of every head are then concatenated and once again projected, resulting in the mLSTM layer
\begin{align} \label{eq:xlstm_layer_multihead}
    \hspace{-0.3em} 
    \mathrm{mLSTM}(\BX) &= \mathrm{Concat}(\BH^{(1)}, \dots, \BH^{(N_\text{head})}) \ \BW^{\top}_\text{proj}, 
\end{align}
where $\BH^{(i)} = \mathrm{mLSTMCell}^{(i)}(\BX)$.
We discuss key considerations for choosing the number of parallel heads or in other words the head dimension $d_{hv}$ in Sec.~\ref{sec:xlstm_arch_efficiency}.

\section{Optimized \xlstmlarge Architecture}\label{sec:method}

\begin{figure}
    \centering
    \includegraphics[width=0.47\textwidth]{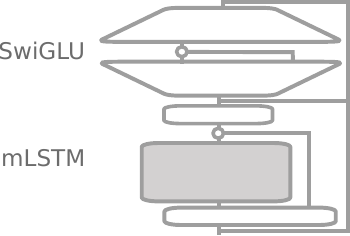}
    \vspace{-0.8em}
    \caption{Sketch of the updated xLSTM Block. 
    The lower part is an output-gated sequence-mix layer with the mLSTM at its core, whereas the upper part is a gated MLP (SwiGLU) as a feature/channel-mix layer. 
    See Fig.~\ref{fig:mLSTMblock_detail} for details.
    \label{fig:mLSTMblock_v1}}
    \vspace{-0.4cm}
\end{figure}

The emerging paradigm of increasing test-time computation necessitates i) the development of novel architectures optimized for \emph{efficient inference}. 
Additionally, new architectures must ii) be viable in large-scale pre-training setups, thus be \emph{highly efficient during training}, and iii) exhibit \emph{stable convergence}.
Our \xlstmlarge is designed to meet these three challenges by offering an architecture that can be trained efficiently and with stable convergence and is also highly efficient at inference. 
In Sec.~\ref{sec:xlstm_arch_efficiency}, we detail our optimization of the xLSTM architecture for \emph{efficiency} during both inference and training.
We then describe in Sec.~\ref{sec:xlstm_arch_stability} our actions to improve and ensure \emph{stable convergence} for training large xLSTM models, focusing specifically on the gating mechanism of the mLSTM cell.

\subsection{Optimizing for Efficiency}
\label{sec:xlstm_arch_efficiency}
The core of the \xlstmlarge architecture, the mLSTM cell, with its recurrent and parallel mode enable efficient inference and training. 
To leverage its full potential, we revisit the design of the surrounding block structures.

\vspace{-0.5em}
\paragraph{Previous mLSTM Block.} 
Similarly to other linear RNNs like Mamba~\citep{Gu2024Mamba, Hua:22transformerlineartime}, the previous xLSTM architecture places the mLSTM cell combined with channel-wise convolutions 
in between a linear up-projection and down-projection, which is referred to as \emph{pre up-projection block}~\citep{Beck2024xLSTM}. 
These blocks combine sequence mixing and channel mixing in one block and are therefore stacked homogeneously without interleaving position-wise feed-forward MLP layers.
Although the pre up-projection block architecture has proven competitive language modeling performance for the xLSTM up to 1.4B parameters, it comes with a substantial trade-off in computational efficiency for the following reasons:
\vspace{-0.2cm}
\begin{enumerate}
 \item Within the pre up-projection block, the mLSTM operates in a significantly higher dimension than the embedding dimension of the model. 
This leads to a substantially \emph{higher computational cost and GPU memory usage for the mLSTM operation}.
\item Omitting position-wise feed-forward MLP layers results in a \emph{decreased proportion of highly efficient linear layer FLOPs} in the model.
\item The previous xLSTM architecture uses several additional components such as learnable skip connections, channel-wise convolutions, and small (block-diagonal) projection layers to compute queries, keys and values.
Without custom kernel fusion, these small operations result in multiple short kernel calls on the GPU, which cannot effectively utilize tensor cores\footnote{Tensor cores are specialized compute units that accelerate matrix multiplications on GPUs.}
and, consequently, significantly \emph{reduce GPU utilization}.
\item Previously, the input and forget gate pre-activations were computed from concatenated query, key and value projections. 
In a large-scale tensor-parallel training setup this requires an additional all-reduce operation per mLSTM block, which \emph{increases the overall communication cost}.
\end{enumerate}
\vspace{-0.2cm}

These limitations prevent efficient scaling of the xLSTM architecture as introduced by~\citet{Beck2024xLSTM} beyond 1.4B parameters.
To scale the xLSTM to even larger model sizes, we optimize the mLSTM block for maximal efficiency by addressing these four limitations.

\vspace{-0.5em}
\paragraph{Optimizing the mLSTM Block.}

To begin, we operate the mLSTM cell in the models' embedding dimension, instead of a higher dimensional space and place position-wise feed-forward MLP layers after each mLSTM layer.
This modification increases the proportion of highly optimized linear layer (i.e.\ matrix multiplication) FLOPs and reduces the computation cost of the mLSTM operation (see App.~\ref{app:flop_counting} for details on the FLOP computation). 
The significantly reduced GPU memory usage enables larger batch sizes during training, which also increases training efficiency.
The result is the default dense Transformer block configuration referred to as \emph{post up-projection block} by~\citet{Beck2024xLSTM}:
\vspace{-0.1cm}
\begin{subequations} \label{eq:xlstm_block}
\vspace{-0.7em}
\begin{align}
    \bm{z} &= \bm{x} + \mathrm{mLSTM}\big(\mathrm{Norm}(\bm{x})\big), \\
    \bm{y} &= \bm{z} + \mathrm{MLP}\big(\mathrm{Norm}(\bm{z}) \big),
\end{align}
\end{subequations}
where $\bm{x}$ is the input to the block, $\bm{z}$ is the intermediate output of the mLSTM layer defined in~\eqref{eq:xlstm_layer_multihead}, and $\bm{y}$ is the block output. 
The MLP is a SwiGLU \cite{Shazeer2020GLUvariants} (see Fig.~\ref{fig:mLSTMblock_v1}).

Moreover, we discard operations like the channel-wise convolution and the learnable skip-connection, and replace the block-wise query, key and value projections by dense linear layers. 
This again increases linear layer FLOPs and ensures effective usage of tensor cores within the mLSTM layer. 

Finally, we ensure that the gate pre-activations for every head are computed independently as outlined in \eqref{eq:xlstm_layer_multihead}. 
This allows us to apply the model parallelization strategies optimized for Transformers with self-attention~\citep{Shoeybi2020MegatronLM} to our \xlstmlarge architecture and therefore minimize additional communication cost.

These optimizations result in our optimized mLSTM block described in Fig.~\ref{fig:mLSTMblock_v1} and Fig.~\ref{fig:mLSTMblock_detail} in the appendix, of which we stack 32 in our \xlstmlarge architecture.
We observe that our optimizations achieve a 3.5× speedup in training for 1.4B models, with a slight trade-off in validation perplexity that can be mitigated by %
a few more training steps (see Tab.~\ref{tab:v1_vs_v2}).
Although the modified block structure reduces the size of the mLSTM cell memory states $\BC$, we find that it does not compromise the language modeling quality of our model.

\vspace{-0.5em}
\paragraph{Optimizing the Memory Capacity.}
The overall memory capacity of the xLSTM, i.e.\ the amount of information that can be stored from an input sequence, is related to the physical size of its memory cell states $\BC$ of shape $d_{qk} \times d_{hv}$ in GPU memory. 
By choosing either the number of heads or the head dimension $d_{hv}$, the other is given by the relation to the embedding dimension $d=\text{\#heads} \times d_{hv}$.
For the \xlstmlarge we set $d_{qk} = d_{hv}/2$ similar to~\citet{Sun2023RetNet}.
We can then compute the total memory state size by $\text{\#blocks} \times \text{\#heads} \times d_{qk} \times d_{hv} \times 4 \, \text{bytes}$, assuming that the state is stored in \texttt{float32} format. 
In Tab.~\ref{tab:abl_head_dim_7B} we show the memory state size for different number of heads as well as their trade-offs with language modeling performance and training efficiency. 
We use a larger memory state size and a slightly longer train step time to make sure the model is not constrained by a lack of memory.
We elaborate further on this in Sec.~\ref{sec:experiments}. 
We choose 8 heads with head dimension $d_{hv}=512 \,$ for \xlstmlarge. 

\vspace{-0.5em}
\paragraph{Fused Generation Kernels for the mLSTM Cell.}
During autoregressive generation, the hidden state outputs of the mLSTM cell are computed, with its recurrent formulation given by \eqref{eq:mlstm_rec}~--~\eqref{eq:mlstm_rec_end}. 
The recurrent formulation consists of a combination of an outer-product, dot-products and several pointwise operations, which translates to individual consecutive GPU kernels. 
Since each kernel loads its inputs from and stores its outputs to GPU memory, this increases the amount of slow memory operations. 
To ensure that intermediate results of equations~\eqref{eq:mlstm_rec_begin}--\eqref{eq:mlstm_rec_htilde} are not unnecessarily transferred to GPU memory, but instead remain on the GPU's compute chips, we write fused GPU kernels for the mLSTM generation mode. 
This results in significantly faster generation as shown in speed benchmarks in Sec.~\ref{sec:experiments:speedbenchmarks}.

\begin{figure*}
    \centering
    \includegraphics[width=\textwidth]{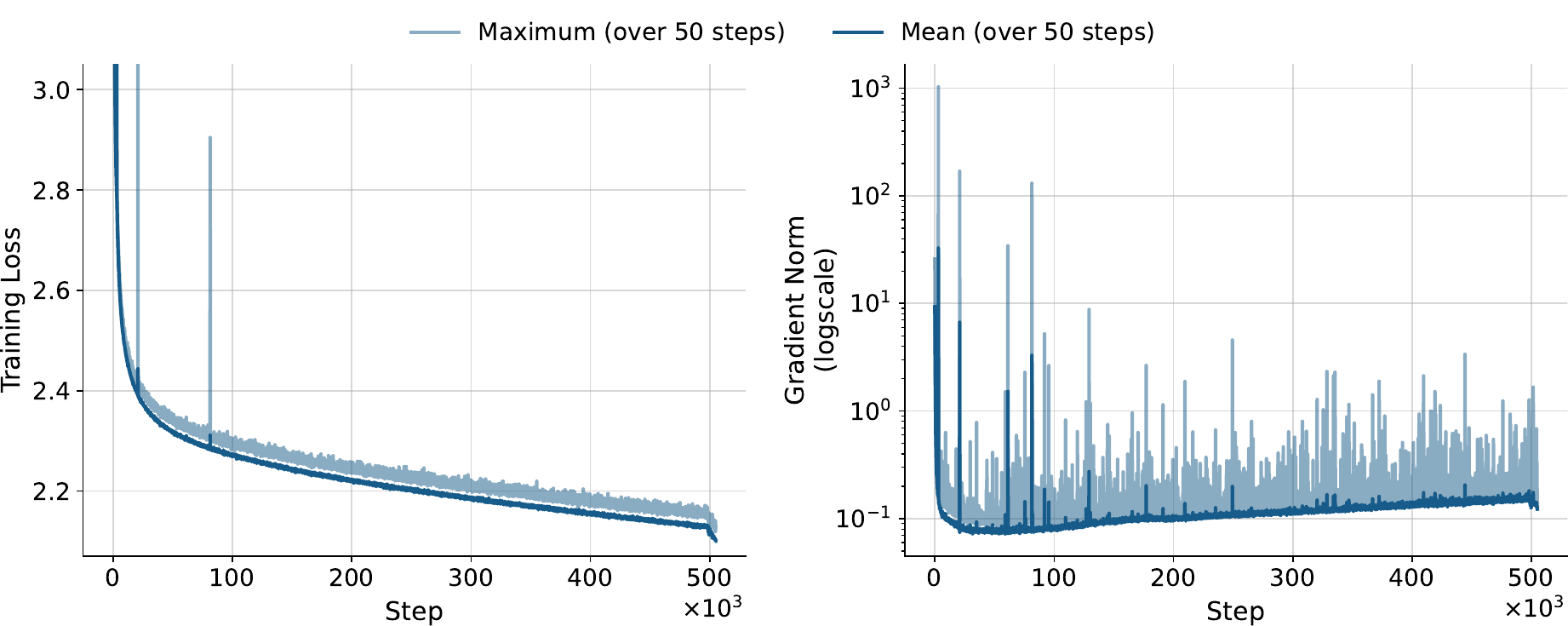}
    \vspace{-0.8em}
    \caption{Loss and Gradient Norm during Pretraining of xLSTM 7B.
    We show the mean and maximum value over 50 steps. 
    Our enhanced architecture and initialization enable stable pretraining of xLSTM 7B, exhibiting only two brief loss spikes early in training, both of which were rapidly recovered. 
    \label{fig:loss_curve}}
    \vspace{-.3em}
\end{figure*}

\begin{table*}
\caption{Model Performance on Huggingface Leaderboard v2. $\uparrow$ indicates larger values are better.}
\vspace{0.2cm}
\label{tab:leaderboardv2}
    \begin{center} \begin{small}
    \begin{tabular}{lccccccc}
        \toprule
        \textsc{Model} & \textsc{BBH} $\uparrow$ & \textsc{MMLU-Pro} $\uparrow$ & \textsc{Math} $\uparrow$ & \textsc{MuSR} $\uparrow$ & \textsc{GPQA} $\uparrow$ & \textsc{IFEval} $\uparrow$ & \textsc{Average} $\uparrow$ \\ 
        \midrule \textsc{Transformers} &&&&&&& \\ 
        Llama-3.1-8B & 0.465 & 0.325 & 0.042 & 0.379 & 0.312 & 0.125 & 0.275 \\ 
        Llama-2-7B-hf & 0.349 & 0.186 & 0.013 & 0.363 & 0.269 & 0.264 & 0.241 \\ 
        OLMo-7B-hf & 0.330 & 0.118 & 0.010 & 0.357 & 0.257 & 0.280 & 0.225 \\ 
        Gemma-7B & 0.426 & 0.293 & 0.061 & 0.408 & 0.295 & 0.272 & 0.292 \\ 
        Ministral-8B-Instruct-2410 & 0.496 & 0.350 & 0.151 & 0.430 & 0.319 & 0.322 & 0.345 \\ 
        Bloom-7B1 & 0.311 & 0.111 & 0.000 & 0.354 & 0.264 & 0.138 & 0.196 \\ 
        Gpt-j-6B & 0.321 & 0.125 & 0.009 & 0.363 & 0.261 & 0.250 & 0.222 \\ 
        Pythia-6.9B & 0.326 & 0.116 & 0.006 & 0.355 & 0.270 & 0.232 & 0.217 \\ 
        Qwen2.5-7B & 0.541 & 0.435 & 0.165 & 0.446 & 0.329 & 0.359 & 0.379 \\ 
        Gemma-2-9B & 0.543 & 0.414 & 0.117 & 0.453 & 0.334 & 0.217 & 0.346 \\ 
        DCLM-7B & 0.426 & 0.312 & 0.030 & 0.392 & 0.303 & 0.228 & 0.282 \\ 
        \midrule \textsc{Transformer-Recurrent hybrids} &&&&&&& \\ 
        Zamba2-7B & 0.489 & 0.319 & 0.114 & 0.402 & 0.318 & 0.375 & 0.336 \\ 
        \midrule \textsc{Recurrent models} &&&&&&& \\ 
        Falcon-Mamba-7B (pre-decay) & 0.373 & 0.177 & 0.024 & 0.387 & 0.275 & 0.252 & 0.248 \\ 
        Falcon-Mamba-7B & 0.429 & 0.229 & 0.039 & 0.412 & 0.299 & 0.335 & 0.290 \\ 
        MambaCodestral-7B (v0.1) & 0.405 & 0.191 & 0.023 & 0.359 & 0.266 & 0.322 & 0.261 \\ 
        RKWV-v5-Eagle-7B & 0.325 & 0.121 & 0.007 & 0.322 & 0.243 & 0.266 & 0.214 \\
        RWKV-v6-Finch-7B & 0.342 & 0.154 & 0.014 & 0.338 & 0.265 & 0.264 & 0.230 \\
        \textbf{\xlstmlarge}               & 0.381 & 0.242 & 0.036 & 0.379 & 0.280 & 0.244 & 0.260 \\
        \textbf{\xlstmlarge} LCTX & 0.390 & 0.252 & 0.040 & 0.374 & 0.253 & 0.234 & 0.257 \\
        \bottomrule
    \end{tabular}
    \end{small} \end{center} 
    \vspace{-0.4cm}
\end{table*}

\subsection{Optimizing for Stability}
\label{sec:xlstm_arch_stability}
We find that the previous xLSTM architecture at the 7B parameter scale often becomes unstable in early stages of training.
In particular, we noticed that training at higher learning rates leads to large spikes in the gradient magnitude and loss value, similar to reports from previous works on Mamba-based models \cite{Lieber2024Jamba, Dao2024Mamba2, Buo2024FalconMamba}.
We further observed and attribute these spikes to very large outlier features, i.e.\ individual feature values that are significantly larger than the average feature value \cite{He2024OutlierFeatures}. 
We address these stability issues by (i) the use of RMSNorm instead of LayerNorm, (ii) soft-capping of the input and forget gates, and (iii) a negative initialization of the input gate bias. 
\vspace{-0.5em}
\paragraph{Pre-Norm with RMSNorm.}
Many works report that replacing the LayerNorm by RMSNorm at the input of each layer (e.g.\ in the pre-norm setting~\citep{Xiong2020OnLayerNormalizationTransformer}) improves training stability for Transformers~\citep{olmo2025OLMo2, Meta2023Llama2, Google2024Gemma, Yang2024Qwen2} and Mamba models~\citep{Buo2024FalconMamba}.
Our experiments in App.~\ref{app:ablation_experiments}, Fig.~\ref{fig:norm_comparison} confirm that this also applies to the \emph{pre-norm} normalization layers in \eqref{eq:xlstm_block} in our xLSTM architecture.
Therefore, we replace the LayerNorm by RMSNorm in our xLSTM architecture. 

\vspace{-0.5em}
\paragraph{Gate Soft-Capping.}
To reduce potential large outlier features and related loss spikes, we apply soft-capping to the input and forget gate pre-activations $\tilde{\Ri}_t$ and $\tilde{\Rf}_t$, such that their values stay between $-a$ and $a$ for a specific cap value $a$. 
We cap the gates using $a=15$ with the function
\begin{equation}
    \label{eq:softcap}
    \mathrm{softcap}_a(\Bx) = a \cdot \tanh(\Bx / a).
\end{equation}
In Sec.~\ref{sec:experiments:ablations} and App. Sec.~\ref{app:ablation_experiments}, we confirm that this significantly improves the stability and performance of our xLSTM architecture.
Additionally, we apply soft-capping with $a=30$ to the final layer logits, similar to~\citet{gemmateam:24gemma2}.

\vspace{-0.5em}
\paragraph{Negative Input Gate Bias Initialization.}
\label{sec:input_gate_bias_init_desc}
We observe that early on in training our xLSTM models experience large gradient norm spikes, which affect the final performance of our model (see Fig.~\ref{fig:negative_input_gate_bias_init} in App.~\ref{app:ablation_experiments}). 
Initializing the input gate at large negative values (e.g. -10) effectively mitigates these gradient norm spikes and improves performance. 
We analyze the impact of the input gate further in Sec.~\ref{sec:experiments:ablations}.

In summary, our optimizations enable remarkably stable pretraining of xLSTM 7B, as we show in Figure~\ref{fig:loss_curve}.

We outline the detailed block architecture of our \xlstmlarge in Appendix~\ref{app:block_architecture} and our training recipe in Appendix~\ref{sec:method:training}.

\vspace{-0.1cm}
\section{Related Work}\label{sec:relatedwork}
\vspace{-0.1cm}

Although the largest language models to date have predominantly relied on Transformer-based architectures, recurrent LLMs and hybrid models have recently gained traction as alternative architectures due to their enhanced efficiency in processing long contexts. 
Many recent efforts have targeted the 7B parameter scale (or nearby), striking a balance between model capacity and resource constraints.
Griffin \cite{De2024Griffin} is one of the first hybrid recurrent models that was trained with up to 14B parameters. 
Later, the same architecture was used to train RecurrentGemma with 9B parameters \cite{Botev2024RecurrentGemma}. 
The Griffin architecture uses a 1D temporal convolution of size 4 before the sequence mixing part, similar to H3 \cite{Fu2023H3} and Mamba \cite{Gu2024Mamba}, but the hidden state is vector valued with independent updates per each (scalar) dimension.
In contrast, Eagle-7B \cite{Peng2024RWKV6} builds on the RWKV architecture and uses a matrix-valued hidden state similar to linear attention and gated linear attention \cite{Katharopoulos2020LinearAttention, Yang2024GLA}.

Among the Mamba models at the 7B parameter scale, \citet{Waleffe2024MambaStudy} provided the first comparative analysis of Mamba 1, Mamba 2, and a hybrid Mamba architecture. 
In their experiments, the performance of both Mamba 1 and Mamba 2 significantly lagged behind Transformers, while the hybrid architecture was shown to surpass the performance of Transformers. 
Aligned with this finding, several new hybrid Mamba architectures have been proposed, including Samba (3.8B) \cite{Ren2024Samba}, Zamba (7B) \cite{Glorioso2024Zamba}, and the 12B parameter mixture-of-experts-model Jamba \cite{Lieber2024Jamba}. 
More recently, FalconMamba \cite{Buo2024FalconMamba} based on Mamba~1 and Codestral Mamba~\citep{Mistral2024CodestralMamba} based on Mamba~2 have shown that a purely recurrent architecture is capable of exceeding the performance of both hybrid Mamba models and Transformers. 

\vspace{-0.2cm}
\section{Experiments}\label{sec:experiments}
\subsection{Language Modeling Performance}
\paragraph{Huggingface Leaderboard.}
We start by benchmarking \xlstmlarge against state-of-the-art Transformer and recurrent LLMs on the 7B parameter scale. 
To this end, we evaluate the performance on the Open LLM Leaderboard v2 using the LM Evaluation Harness~\citep{eval-harness,open-llm-leaderboard-v2}. 
The results are summarized in Tab.~\ref{tab:leaderboardv2}, showing that \xlstmlarge ranks in the mid-range among 7B-scale models, several of which benefited from substantially larger training datasets. 
We believe that with a larger and better curated training dataset, including a greater emphasis on math and code data in earlier training phases, \xlstmlarge could match the performance of the strongest 7B models.

\vspace{-0.2cm}
\paragraph{Long-Context Evaluation and Fine-Tuning.} \label{sec:long_context_evaluation}
To evaluate long-context capabilities, we use the RULER benchmark \cite{hsieh2024ruler}, which consists of a set of synthetic needle-in-a-haystack, question-answering and variable tracking tasks, with varying context length from 4K to 131K tokens. 
For this benchmark, we consider both our standard \xlstmlarge and a long-context version (\xlstmlarge LCTX), where we replace the standard cool-down phase described in App.~\ref{sec:method:training} with a long-context variant. 
For the long-context cool-down phase, we add long-context data (see App.\ Tab.~\ref{tab:dataset_longctx}) to the training corpus and train the model with a context length of 32K, while adjusting the batch size to maintain the number of tokens per batch.
We compare to Llama 2 7B (not long-context fine-tuned) and Llama 3.1 8B (long-context fine-tuned up to 131K tokens) as Transformer baselines, CodestralMamba and FalconMamba as State Space Model baselines, and RWKV-5/6 as additional RNN baselines.

\begin{figure}[t]
    \centering
    \includegraphics[width=\columnwidth]{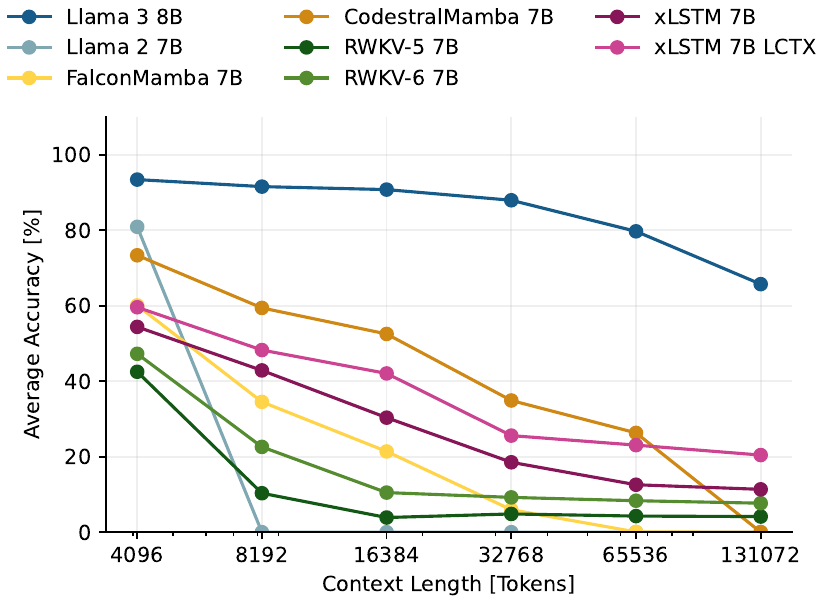}
    \vspace{-0.65em}
    \caption{RULER results of \xlstmlarge in comparison to Transfomers (with and without long context finetuning) and State Space Models, with and without medium context cooldown. \label{fig:ruler_results}}
    \vspace{-0.3em}
\end{figure}

The results on RULER are shown in Fig.~\ref{fig:ruler_results}. 
As expected, Llama 3 provides the strongest baseline, since it is heavily fine-tuned on very long contexts and with a more advanced and optimized approach \cite{Meta2024Llama3}. 
On the other hand, Llama 2 fails entirely for context lengths beyond 4k, for which it has not been trained. 
For \xlstmlarge, the long-context cool-down stage in pre-training largely improves long-context capabilities, resulting in competitive performance compared to state-space models and outperforming RWKV-5/6. 
Notably, the long-context \xlstmlarge achieves 20\% average accuracy at a context length 131k, although it was trained only with a context length up to 32k during the cool-down phase.
This is particularly remarkable given that, unlike Transformers with a growing KV cache, \xlstmlarge must store information from the entire sequence in a fixed-size memory with limited capacity (see Tab.~\ref{tab:abl_head_dim_7B}). 
We assume that \xlstmlarge's performance could be pushed further by explicitly training on even longer sequences and with a more advanced fine-tuning protocol as it was used in the training of Llama 3~\citep{Meta2024Llama3}. 

In Sec.~\ref{sec:experiments:ablations}, we further investigate the effect of the memory state size and the input gate on the long context capabilities of \xlstmlarge.

\subsection{Speed Benchmarks}\label{sec:experiments:speedbenchmarks}
The constant memory size and linear compute scaling with context length of our xLSTM architecture enable highly efficient generative inference in large scale-inference serving environments as well as local inference running on edge devices.

We focus on the local single user inference setting, which is common when models are deployed on edge devices.
Therefore, we benchmark generative inference with our \xlstmlarge model on a single NVIDIA H100 GPU with batch size 1, unless specified otherwise. 
We compare our \xlstmlarge to Llama 2 and Llama 3 models as Transformer baselines and Falcon Mamba (Mamba 1 architecture) and Codestral Mamba (Mamba 2 architecture) as Mamba baselines. 
We use model implementations from Huggingface transformers library and optimize each with \texttt{torch.compile} \footnote{\tiny\url{https://github.com/huggingface/transformers}}
and PyTorch CUDA Graphs~\citep{Nguyen2021PyTorchCUDAGraph}. 

\begin{figure}
    \centering
    \includegraphics[width=\columnwidth]{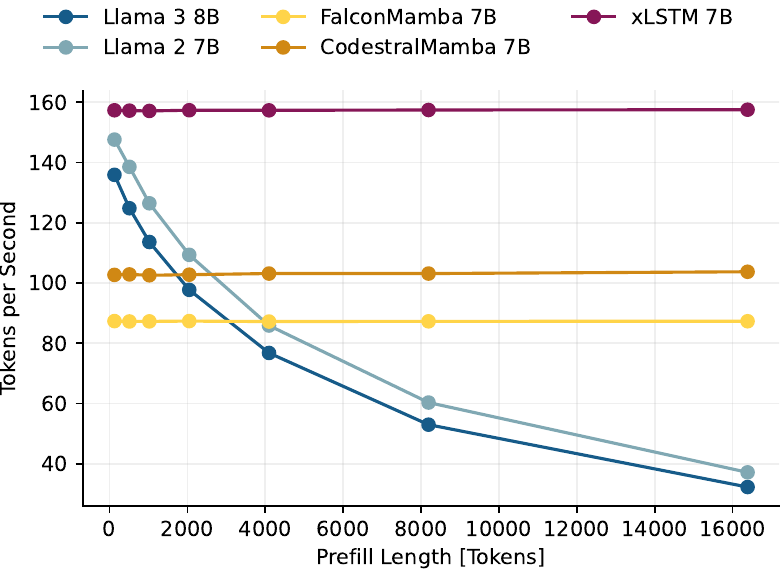}
    \vspace{-0.9em}
    \caption{Throughput for generating 100 tokens with batch size 1 at varying prefill lengths.\label{fig:generation_throughput}}
    \vspace{-.0em}
\end{figure}

\paragraph{Generation Throughput.}
The generation throughput measures the generation speed in tokens per second at varying prefill lengths, i.e., varying length of documents the model gets to read before it starts to generate text.
In Fig.~\ref{fig:generation_throughput}, we observe that due to the quadratic scaling with input context length of the attention mechanism, the speed at which the Transformer models can generate text significantly drops for longer prefill lengths.
In contrast, recurrent architectures with constant cost per generated token have a constant generation speed independent of the input context length.

We find that \xlstmlarge is about 50\% faster in text generation than Mamba, which we attribute mostly to our optimized block design (see Sec.~\ref{sec:method}), and even faster than Llama-based Transformer models with a similar block design at prefill length 0.

\begin{figure}[ht]
    \centering
    \includegraphics[width=\columnwidth]{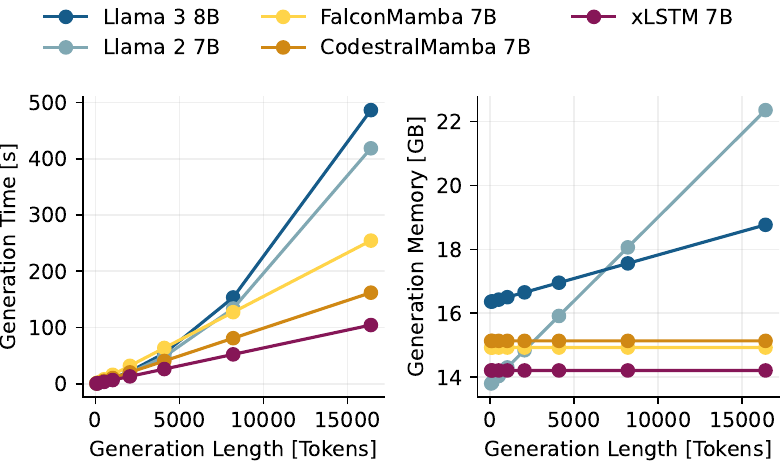}
    \vspace{-0.8em}
    \caption{Time and GPU memory used for generation of a single sequence of varying lengths for generation without prefill. \label{fig:generation_time_and_memory}}
\end{figure}

\paragraph{Generation Time and Memory Consumption.}
We measure the token generation time and GPU memory usage (without pre-fill) for different generation lengths. 
Fig.~\ref{fig:generation_time_and_memory} (left) demonstrates the linear scaling of recurrent models vs.\ the quadratic scaling of Transformers in compute (runtime), while Fig.~\ref{fig:generation_time_and_memory} (right) shows the constant memory size of recurrent models compared to the linear growth of the Transformer KV-cache. 
Since Llama~3 uses grouped query attention~\citep{Ainslie2023gqa} the memory usage grows slower compared to Llama~2, which uses default multi-head attention.

With our optimized block design, we operate the mLSTM in a lower dimensional space. This results in a significantly lower memory footprint (Fig.~\ref{fig:generation_time_and_memory} (right)) and lower generation times (Fig.~\ref{fig:generation_time_and_memory} (left)) of our \xlstmlarge model compared to the Mamba models. 

\begin{figure}[ht]
    \centering
    \includegraphics[width=\columnwidth]{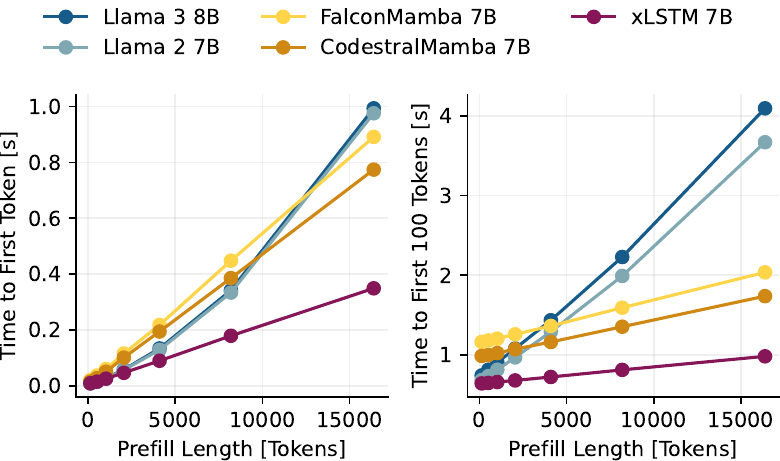}
    \vspace{-0.8em}
    \caption{Time to first (1) token and time to first 100 tokens at varying preﬁll lengths for batch size 1. \label{fig:ttf}}
\end{figure}

\paragraph{Time To First Token.}
In applications, where the language model operates as interface to the user (potentially on edge devices), it is important to have short response times. 
In Fig.~\ref{fig:ttf}, we measure this response time or latency as the time the model takes to generate 1 or 100 token after consuming varying prefill lengths. 
Our \xlstmlarge achieves the fastest response times for all prefill lengths.

\paragraph{Prefill Throughput.}  %
Finally, we measure the prefill throughput in tokens per second for \num{65536} tokens at varying batch size and context length.
Due to the quadratic scaling with context length, the throughput of the Llama models decreases with longer contexts. 
In contrast, our \xlstmlarge achieves the highest throughput (about 70\% higher than Codestral Mamba) independent of the context length.
\begin{figure}
    \centering    \includegraphics[width=\columnwidth]{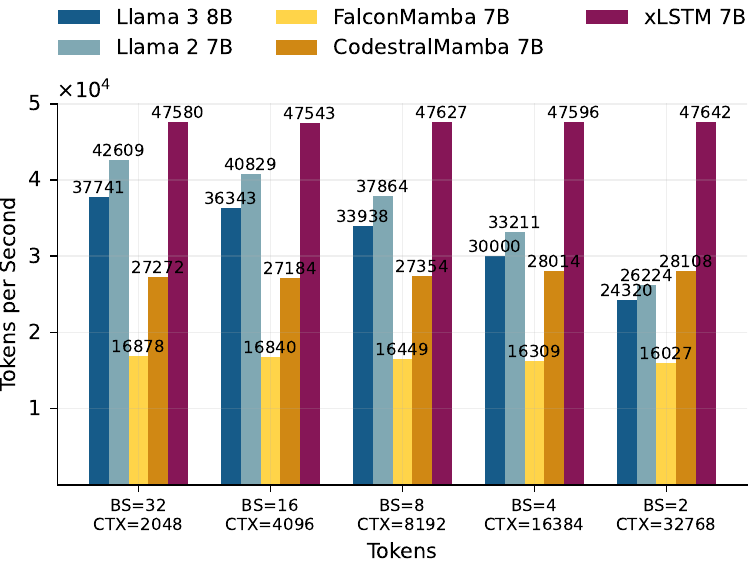}
    \vspace{-1.0em}
    \caption{Prefill throughput varying batch size and context length. \label{fig:throughput}}
\end{figure}

\subsection{Ablation Studies}\label{sec:experiments:ablations}
Finally, we validate our design choices to optimize the training stability and efficiency of our \xlstmlarge architecture.

\paragraph{Pre-Up vs.\ Post-Up Projection Block.}  
We 
compare the pre-up projection block architecture against our optimized mLSTM block in terms of validation perplexity and training step time for three model sizes. 
For both block architectures, we apply gate soft-capping and the input gate bias initialization described in Sec.~\ref{sec:method}.
The results in Tab.~\ref{tab:v1_vs_v2} show only a slight performance difference in terms of validation perplexity at the largest model size. 
However, the $3.5\times$ speedup in training step time confirms our choice for the post-up projection block in \xlstmlarge, deviating from the pre-up projection of Mamba \cite{Gu2024Mamba, Dao2024Mamba2} and the previous xLSTM architecture \cite{Beck2024xLSTM}.

\begin{table}[t]
\caption{Comparison between the previous xLSTM architecture~\citep{Beck2024xLSTM} and our \xlstmlarge architecture in terms of step time and perplexity for different number of parameters.
Models of size 160M and 400M use batch size 128 distributed over 16 GPUs, and 1.4B parameter models use batch size 256 (32 GPUs). For the 7B parameter model, our new architecture uses batch size 512 (128 GPUs), whereas the previous architecture uses only batch size 256 (128 GPUs) because of the architecture's increased GPU memory requirements. 
Due to the expensive computational costs, we only compute the token throughput and did not fully train the 7B parameter models for this ablation. \\
$\uparrow$ / $\downarrow$ indicates larger / smaller values are better.
}
\vspace{-0.35cm}
\label{tab:v1_vs_v2}
\begin{center}
\resizebox{\columnwidth}{!}{%
\begin{small}
\begin{sc}
\begin{tabular}{llcccc}
\toprule
\multicolumn{2}{c}{Model} & Throughput $\uparrow$& Speedup $\uparrow$ & ppl $\downarrow$ & $\Delta$ ppl  \\
                          &  & 1K tokens/sec &    & &      \\
\midrule
160M & Previous & 76.20  &             & 20.43 &      \\
     & Ours & 225.99 & $ \times$2.97 & 21.34 & +0.91 \\
\midrule
400M & Previous  & 28.13 &            & 15.26 &      \\
     & Ours & 102.40 & $\times$3.64 & 15.74 & +0.48 \\
\midrule
1.4B & Previous  & 10.57  &            & 12.46 &      \\
     & Ours & 37.03  & $\times$3.50 & 12.68 & +0.22 \\
\midrule
7B   & Previous  & 3.46   &            & - &      \\
     & Ours  & 9.15  & $\times$ 2.64 &  - &   \\
\bottomrule
\end{tabular}
\end{sc}
\end{small}
}
\end{center}
\vspace{-0.5cm}
\end{table}

\vspace{-1em}
\paragraph{Memory State Size.}
The memory state size as well as the training step time is directly influenced by the number of heads (see Sec.~\ref{sec:xlstm_arch_efficiency} and Tab.~\ref{tab:abl_head_dim_7B}). 
In this experiment we investigate how the memory state size affects the performance of the xLSTM in validation perplexity, on downstream tasks as well as on long context tasks. 
To do so, we train xLSTM models with 7B parameters and different number of heads on 160B tokens of our pre-training dataset. 
In our evaluations in perplexity (Tab.~\ref{tab:abl_head_dim_7B}) and on downstream tasks (Tab.~\ref{tab:leaderboardv2_abl}~and~\ref{tab:leaderboardv1_abl}), we find that the performance remains stable across different the number of heads, i.e., memory state sizes, with a slight improvement for more heads (e.g. 16).
In contrast, our long context evaluation in Fig.~\ref{fig:ruler_head_igate_abl} suggests that at very long contexts 4 and 8 heads (i.e., larger memory states) seem to perform better. 
While this is in line with our intuition that larger memory state size corresponds to better long-context capabilities, we believe that an even larger study (e.g., training on more tokens) than our ablation at 7B parameters and 160B tokens would be necessary to fully explore this connection.

\begin{table}[htbp!]
    \vspace{-0.1cm}
    \caption{Head dimension ablation for a 7B parameter xLSTM model with 32 blocks, embedding dimension 4096 and training context length 8192. \textit{KV Cache in Tokens} shows how many tokens in a similar sized Transformer correspond to our state size. \textit{FLOPs forward} are the mLSTM cell forward FLOPs for a full sequence. $\downarrow$ indicates smaller values are better.}
    \label{tab:abl_head_dim_7B}
    \vspace{0.2cm}
    \begin{adjustbox}{width=0.49\textwidth}    
    \begin{tabular}{cc|cccccc}
    \toprule
    \#Heads & $d_{hv}$ & \thead{Total Memory                               \\ State in MB} & \thead{KV Cache \\ in Tokens} & \thead{FLOPs \\ forward} $\downarrow$ & \thead{Val \\ PPL} $\downarrow$ & \thead{Train Step \\ Time in s} $\downarrow$\\
    \midrule
    4       & 1024     & 268.4               & 256 & 7.6e11 & 9.58 & 3.97 \\
    8       & 512      & 134.2               & 128 & 4.1e10 & 9.52 & 3.63 \\
    16      & 256      & 67.1                & 64  & 2.4e10 & 9.52 & 3.51 \\
    32      & 128      & 33.6                & 32  & 1.5e10 & 9.55 & 3.41 \\
    \bottomrule
\end{tabular}

    \end{adjustbox}
    \vspace{-0.5cm}
\end{table}

\textbf{Norm Layer Types.}
Our update on the xLSTM block architecture has two normalization layers, a pre-norm at the block entry and a head-wise norm layer after the mLSTM cell. 
In this ablation, we test the effect of the types of these normalization layers on training stability and performance, with LayerNorm~\citep{Ba2016LayerNorm} and RMSNorm~\citep{Zhang2019Rmsnorm} as the options. 
In Fig.~\ref{fig:norm_comparison} in App.~\ref{app:ablation_experiments} we confirm that, for the pre-norm the RMSNorm type has a strong stabilizing effect, whereas for the mLSTM cell state norm there is no impact on stability and performance.

\pagebreak
\textbf{Soft-capping.}
Soft-capping (Eq.~\eqref{eq:softcap}) of the output logits and the input and forget gate pre-activations, 
is important for training stability.
In Fig.~\ref{fig:soft_capping} of the appendix, we visualize the validation loss and gradient norms during training on 160B tokens with and without soft-capping. 
The run without soft-capping shows a higher variance in the gradient norms and an overall worse validation loss. 

\textbf{Input Gate.}
We initialize the input gate with larger negative values (e.g. -10) to mitigate large gradient norm spikes and variance (see Sec.~\ref{sec:xlstm_arch_stability}). 
This suggests that the input gate is important for the performance of the xLSTM architecture. 
Therefore, in App.~\ref{app:ablation_experiments} we test the effect of having the input gate non-trainable. 
We compare a version with fixed input gate at one (i.e. setting weights and biases to zero) with a version, where the input gate bias is fixed at our low default initialization value of -10.
We find that, while the learnable input gate only slightly improves performance of our xLSTM over the fixed input gate versions on our standard downstream tasks (App.~\ref{app:ablation_experiments}, Tab.~\ref{tab:leaderboardv2_abl}~and~\ref{tab:leaderboardv1_abl}), it significantly improves performance on long-context evaluations (App.~\ref{app:ablation_experiments}, Fig.~\ref{fig:ruler_head_igate_abl}).

\section{Conclusion}\label{sec:conclusion}
In this work, we demonstrate how our targeted modifications enable the xLSTM architecture to scale to models with 7B parameters, trained on 2.3 T tokens.
By switching to a post-up-projection structure, gate soft-capping and proper initialization, we largely improve training stability and token throughput, making the xLSTM the fastest RNN-based architecture at the 7B scale, while competitive in performance with Transformers and other recurrent models. 
We believe that xLSTM's very high decoding speeds in combination with its good performance highlight its potential as foundational architecture for methods investing substantial compute at inference time.

\section*{Impact Statement}

This paper presents a novel architecture for fast and efficient language modeling, reducing computational costs and energy consumption without sacrificing performance. By making high-quality language models more accessible, our approach helps bridge the digital divide, enabling equitable AI deployment in low-resource settings. Additionally, the efficiency gains contribute to environmental sustainability by lowering the carbon footprint of large-scale NLP systems.
However, there might be both positive and negative societal impacts. We are aware of the risks, but believe that our and the overall advancements in the field of machine learning technology provide a net benefit to society and the world.

\newpage

\bibliographystyle{icml2025}
\bibliography{bibliography}

\newpage
\appendix
\onecolumn

\section{\xlstmlarge Architecture Summary}
\label{app:block_architecture}
The \xlstmlarge architecture consists of 32 post-up projection blocks and is described in Fig.~\ref{fig:mLSTMblock_v1} and Tab.~\ref{tab:model_params}.
We use the GPT-NeoX-20B tokenizer \cite{Black2022GPTNeox} with vocabulary size 50257 and do not tie the weights for input layers (embedding) and output layers (logits). 

\begin{table}[ht]
\caption{Hyperparameters of \xlstmlarge.}
\label{tab:model_params}
\centering
    \vskip 0.15in \begin{center} \begin{small} \begin{sc}
    \begin{tabular}{@{}ccccc@{}}
        \toprule
        \thead{Num\\Params} & \thead{Vocab\\Size} & \thead{Num\\Blocks} & \thead{Model\\Dim} & \thead{Num\\Heads} \\
        \midrule
        6,865,424,896 & 50257 & 32 & 4096 & 8 \\
        \bottomrule
    \end{tabular}
    \end{sc} \end{small} \end{center} \vskip -0.1in
\end{table}

\begin{figure}[H]
    \centering
    \includegraphics[width=0.6\columnwidth]{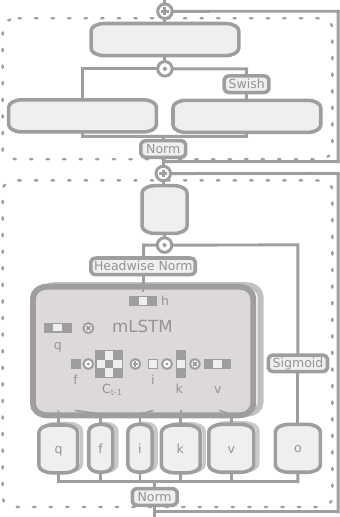}
    \vspace{-0.8em}
    \caption{Improved xLSTM Block. 
    The lower part is a output-gated sequence-mix layer with the mLSTM at its core, whereas the upper part is a Gated MLP (SwiGLU) as a feature/channel-mix layer. 
    Multiple Heads are shown in depth, larger light gray boxes without are linear layers. 
    For the SwiGLU we use a projection factor of 2.66 matching common Transformers. For the query/key dimension we use a factor of 0.5. The Norm layers are RMS norms~\citep{Zhang2019Rmsnorm}, the Headwise Norm is a Layernorm~\citep{Ba2016LayerNorm}.
    \label{fig:mLSTMblock_detail}}
\end{figure}

\pagebreak
\section{Training Recipe}\label{sec:method:training}
\paragraph{Optimization.}
Pre-training was conducted on a high-performance computing cluster comprising 128 NVIDIA H100 GPUs.
We use Fully Sharded Data Parallel (FSDP) and activation checkpointing to reduce the parameter and activation memory footprint. 
We pre-train \xlstmlarge for a total of $550$K (thousand) training steps with batch size $512$ and context length $8192$, encompassing a total of $2.3$T (trillion) training tokens. 
We apply batch size ramp-up with batch size $128$ for the first \num{2000} steps, $256$ for the next \num{2000} steps, and the full batch size ($512$) afterward. 
We use the AdamW optimizer \cite{Loshchilov2018AdamW} with (peak) $\alpha=5\times10^{-4}$, $\beta_1=0.99$, $\beta_2=0.95$, $\epsilon=10^{-8}$, weight decay $0.1$ and gradient clipping norm $0.5$. 
The learning rate schedule comprises a linear warm-up over \num{3000} training steps, an exponential decay phase that spans \num{540000} steps, and a linear cool-down lasting \num{7000} steps.
The exponential decay factor is chosen so that $0.1 \times \alpha$ is reached after \num{500000} steps.

\paragraph{Sequence packing.}
Language datasets come with documents of highly varying lengths. 
To efficiently train a model by processing fixed sequence length sequences (e.g. 8192 tokens), multiple shorter documents are typically packed into a sequence, and the different documents are separated by an end-of-document (EOD) token. 
In order to avoid leaking information between independent documents that are packed into the same sequence, we reset the memory states of each mLSTM cell at the document borders signified by the EOD token. 
This can be easily achieved by explicitly setting the forget gate value to zero, resetting the memory state to the zero matrix.

\paragraph{Dataset selection.}
\label{sec:dataset_cooldown}
We only use publicly available high-quality datasets for pre-training. 
The dataset selection is divided into two training stages: 
In the first stage lasting $500$K (thousand) training steps, we train exclusively on the DCLM dataset \cite{Li2024DCLM}. 
In the second stage ($50$K steps) towards the end of the training, we use a combination of datasets that prioritizes math, coding, and question-and-answer (Q\&A) data. 
The dataset proportions for the second stage are listed in the second column of Tab.~\ref{tab:dataset_longctx}.

Similarly to \citet{Buo2024FalconMamba}, the second training stage includes a collection of small supervised fine-tuning (SFT) Q\&A datasets to improve the model's understanding of texts involving questions and answers. 
These SFT datasets are all publicly available and consist of NuminaMath CoT \cite{NuminaMathDatasets}, MetaMathQA \cite{Yu2023MetaMath}, Tulu v3.1 \cite{Lambert2024Tulu3}, OpenHermes 2.5 \cite{OpenHermes25}, GSM8K \cite{Cobbe2021GSM8K}, and Smoltalk (subsets magpie-ultra, longalign, and self-oss-instruct) \cite{Allal2024SmolLM2}.

For longer context training we replace the high-quality data cool-down by a longer context version keeping the number of tokens per step and the number of steps fixed. The batch size is reduced from 512 to 128, while increasing the context length to 32768.
We replace a large share of the DCLM dataset part with long context text collections, namely LongDataCollections~\citep{togethercompute_longdatacollections_2023}, LongAlign10k~\citep{bai2024longalign},~AntiHayStack~\citep{pan_belandros_anti-haystack_2024} and~LongAlpaca12k~\citep{longlora}, see third column of~Tab.~\ref{tab:dataset_longctx}.

\begin{table}[ht]
    \centering
    \caption{Dataset Proportions for second training stage in standard and longer context mode.}
    \label{tab:dataset_longctx}
    \vskip 0.15in \begin{center} \begin{small}
    \begin{tabular}{lcc}
        \toprule
        \textsc{Dataset Name} & \textsc{Proportion Standard} & \textsc{Proportion LongCtx} \\
        \midrule
        DCLM \cite{Li2024DCLM} & 40\% & 20 \%\\
        FineWeb-Edu \cite{Lozhkov2024FinewebEdu} & 15\% & 15\% \\
        Cosmopedia \cite{Benallal2024Cosmopedia} & 10\%  & 10\%\\
        ProofPile-2 \cite{Azerbayev2023Llemma} & 15\% & 15\% \\
        TheStack \cite{Kocetkov2023TheStack} & 15\%& 15\% \\
        SFT datasets (see Sec.~\ref{sec:method:training})& 5\% & 5\% \\
        LongDataCollections \cite{togethercompute_longdatacollections_2023} & - & 15\% \\
        LongAlign10k \cite{bai2024longalign} &  - & 1\%  \\
        AntiHayStack \cite{pan_belandros_anti-haystack_2024} & - & 1\%  \\
        LongAlpaca12k \cite{longlora}& - & 2\% \\
        \bottomrule
    \end{tabular}
    \end{small} \end{center} \vskip -0.1in
\end{table}

\paragraph{Ablation Training}
\label{app:ablation_training_desc}
For hyperparameter tuning and ablation trainings ("-abl") at the 7B scale, we use a shorter training cycle with 
\num{76000} training steps at context length \num{8192} and batch size 256, resulting in 160B tokens. We use a linear warmup of \num{3000} steps, cosine decay to 10\% of the peak learning rate at \num{75000} steps and a linear cooldown of 
1,000 steps to learning rate 0 at the end. Here, we only train on a subset of the DCLM dataset, without high-quality data in the late phase of pre-training. Peak learning rate and other training hyperparameters are the same as for the main training.

\newpage
\section{Experiments}
\subsection{Extended Evaluation}
To enable comparability to older models, we evaluate our models on the task selection from the first version of the HuggingFace leaderboard using HuggingFace's lighteval~\citep{open-llm-leaderboard,Fourrier2023Lightevalv1}. The results in Tab.~\ref{tab:leaderboardv1} show that there is a trend upwards in metrics from older (e.g. Llama 2) to newer models (e.g. Llama 3.1), but that the differences and ordering between models vary across the tasks.
\begin{table*}[h]
\caption{Model Performance on Huggingface Leaderboard v1 based on {lighteval} by HuggingFace. $\uparrow$ indicates larger values are better.}
\label{tab:leaderboardv1}
\vspace{0.2cm}
    \begin{adjustbox}{width=\textwidth}    
    \begin{tabular}{lcccccccc}
        \toprule
        \textsc{Model} & \textsc{ARC-C} $\uparrow$& \textsc{MMLU} $\uparrow$& \textsc{HellaSwag} $\uparrow$& \textsc{Winogrande} $\uparrow$ & \textsc{TruthfulQA} $\uparrow$& \textsc{OpenBookQA} $\uparrow$ & \textsc{PiQA} $\uparrow$& \textsc{Average} $\uparrow$\\ 
        \midrule \textsc{Transformers} &&&&&&&& \\ 
        
Llama-3.1-8B & 0.562 & 0.663 & 0.720 & 0.745 & 0.362 & 0.447 & 0.818 & 0.617\\ 
Llama-2-7B-hf & 0.511 & 0.468 & 0.687 & 0.706 & 0.318 & 0.412 & 0.786 & 0.555\\ 
OLMo-7B-hf & 0.443 & 0.286 & 0.673 & 0.661 & 0.301 & 0.383 & 0.801 & 0.507\\ 
Qwen2.5-7B & 0.617 & 0.753 & 0.700 & 0.717 & 0.478 & 0.458 & 0.804 & 0.647\\ 
Gemma-7B & 0.593 & 0.640 & 0.721 & 0.740 & 0.381 & 0.436 & 0.813 & 0.618\\ 

\midrule \textsc{Hybrid Models} &&&&&&&& \\
Zamba2-7B & 0.672 & 0.683 & 0.740 & 0.801 & 0.479 & 0.468 & 0.802 & 0.664\\ 

\midrule \textsc{Recurrent Models} &&&&&&&& \\
Falcon-Mamba-7B & 0.599 & 0.622 & 0.709 & 0.743 & 0.459 & 0.460 & 0.822 & 0.631\\ 
Falcon-Mamba-7B (pre-decay) & 0.520 & 0.573 & 0.699 & 0.719 & 0.312 & 0.430 & 0.801 & 0.579\\ 
Mamba-Codestral-7B (v0.1)	& 0.486 & 0.501	& 0.626	& 0.618	& 0.358 & 0.380 & 0.771 & 0.534 \\
RWKV-v5-Eagle-7B & 0.449 & 0.313 & 0.622 & 0.663 & 0.330 & 0.393 & 0.772 & 0.506\\ 
RWKV-v6-Finch-7B & 0.471 & 0.442 & 0.656 & 0.696 & 0.347 & 0.399 & 0.792 & 0.543\\ 
\textbf{\xlstmlarge}  & 0.574 & 0.578 & 0.714 & 0.738 & 0.419 & 0.448 & 0.819 & 0.613\\ 
\textbf{\xlstmlarge} LCTX & 0.516 & 0.588 & 0.715 & 0.740 & 0.374 & 0.429 & 0.819 & 0.597\\ 
        \bottomrule
    \end{tabular}
\end{adjustbox}
\end{table*}

\subsection{Ablation Experiments}
\label{app:ablation_experiments}
\begin{figure}[H]
    \centering
    \includegraphics[width=\columnwidth]{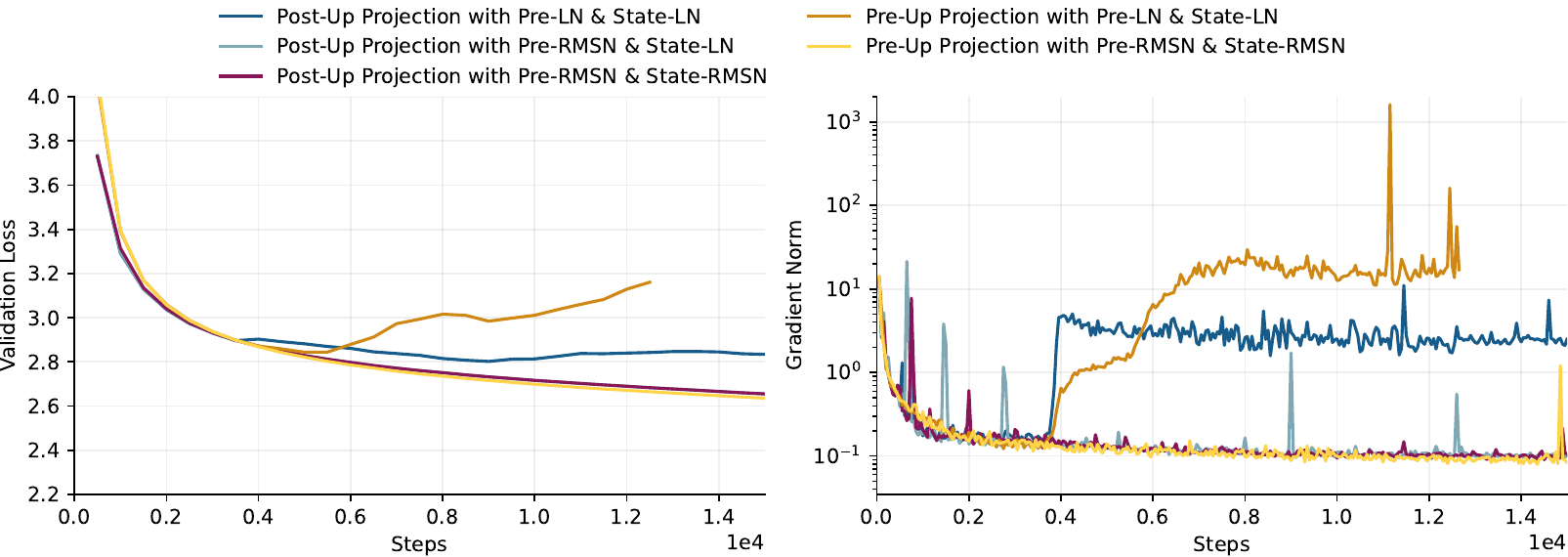}
    \vspace{-0.8em}
    \caption{Comparison of pre-up projection and
    post-up projection blocks with different
    combinations of RMSNorm and LayerNorm.
    At each step, the plot shows the maximum gradient norm observed within the previous 50 steps.
    ~\label{fig:norm_comparison}}
\end{figure}

\paragraph{Effect of the Pre-norm Layer Choice (Fig.~\ref{fig:norm_comparison}).}

Here we asses the effect of different normalization layer choices for the pre-norm in \eqref{eq:xlstm_block} and the state-norm in \eqref{eq:mlstm_rec_hidden_state_output}, both for the xLSTM with a pre-up projection block of \citet{Beck2024xLSTM} and our new post-up projection architecture used for \xlstmlarge. 
We use soft-capping and the negative input bias initialization (see Sec.~\ref{sec:xlstm_arch_stability} and \ref{sec:experiments:ablations}) for both architectures. 
For this experiment, we train models with 1.4B parameters for \num{31000} steps using context length $8192$ and batch size $256$.
Fig.~\ref{fig:norm_comparison} shows the validation loss and gradient norm for the different architectures and normalization layer choices over the course of training (only the \num{15000} steps are shown). 
As can be seen, using LayerNorm as the pre-norm layer leads to very large gradient norms and diverging validation loss after a few training steps, whereas models with RMSNorm train stably. 
For the state-norm layer, the norm type has no impact on the training dynamics. 

\begin{figure}[h]
    \centering
    \includegraphics[width=\columnwidth]{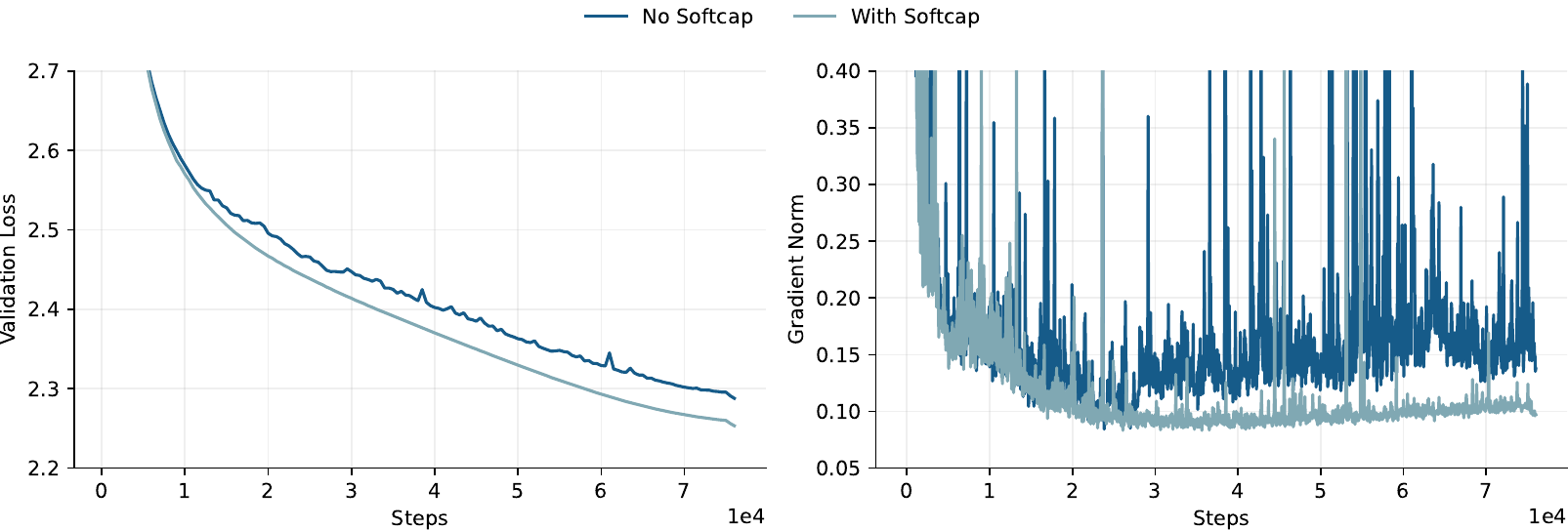}
    \vspace{-0.8em}
    \caption{Effect of softcapping. Two 7B sized xLSTM models are trained with and without soft-capping for 160B tokens. The lower gradient norm noise on the right is a clear indicator for better model performance on the left of the model trained with softcapping. 
    At each step, the plot shows the maximum gradient norm observed within the previous 50 steps.
    ~\label{fig:soft_capping}}
\end{figure}

\paragraph{Effect of Soft-Capping (Fig.~\ref{fig:soft_capping}).}
The two runs in Fig.~\ref{fig:soft_capping} show the effect of soft-capping for two 7B sized xLSTM models trained for \num{76000} steps at batch size 256 and context length \num{8192}, for an effective 160B tokens.

\begin{figure}[H]
    \centering
    \includegraphics[width=\columnwidth]{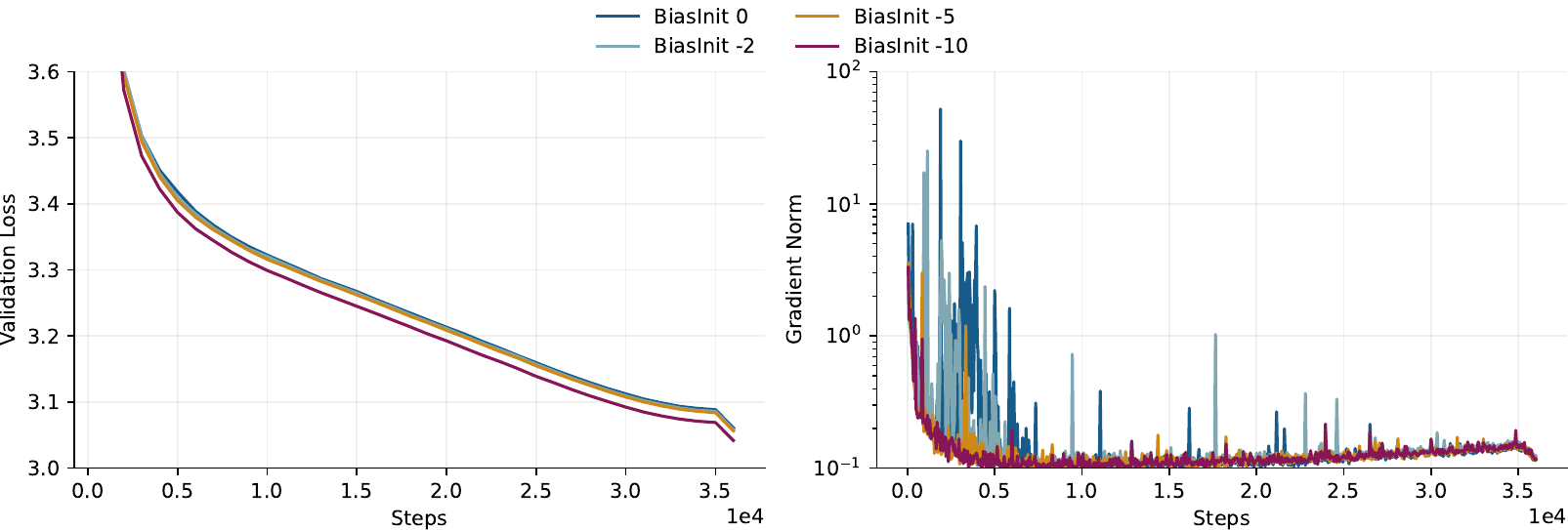}
    \vspace{-0.8em}
        \caption{
    \label{fig:negative_input_gate_bias_init}
    Effect of the Bias Initialization. We conduct experiments with four different input gate biases at the 160M parameter scale, with validation loss on depicted to left and gradient norm on the right, along the training steps. The higher input gate bias initializations show large gradient norm spikes, which results in worse training results. Only the lowest initialization can maintain smooth and low gradient norms with at the best validation perplexities. The reason for this behavior is studied in more detail in~\citep{Anonymous2025TiledFlashLinearAttention}. 
    At each step, the plot shows the maximum gradient norm observed within the previous 50 steps.
    }
\end{figure}

\paragraph{Effect of Negative Input Gate Bias Init (Fig.~\ref{fig:negative_input_gate_bias_init}).}
In this experiment we train 160M parameter models with batch size 128 and context length 4096 and vary the input gate bias initialization [0, -2, -5, -10]. 
The weights of the input gates are initialized to 0. 

In Figure~\ref{fig:negative_input_gate_bias_init} we observe that initializing the input gate biases at -10 effectively mitigates gradient norm spikes and reduces gradient norm variance during training. 
In our experiments up to 7B parameters we observed this behavior transfers across model scales. 

We therefore initialize the input gate biases to -10.
For an extensive discussion of this behavior we refer to concurrent work by~\citet{Anonymous2025TiledFlashLinearAttention}.

\begin{figure}[H]
    \centering
    \includegraphics[width=\columnwidth]{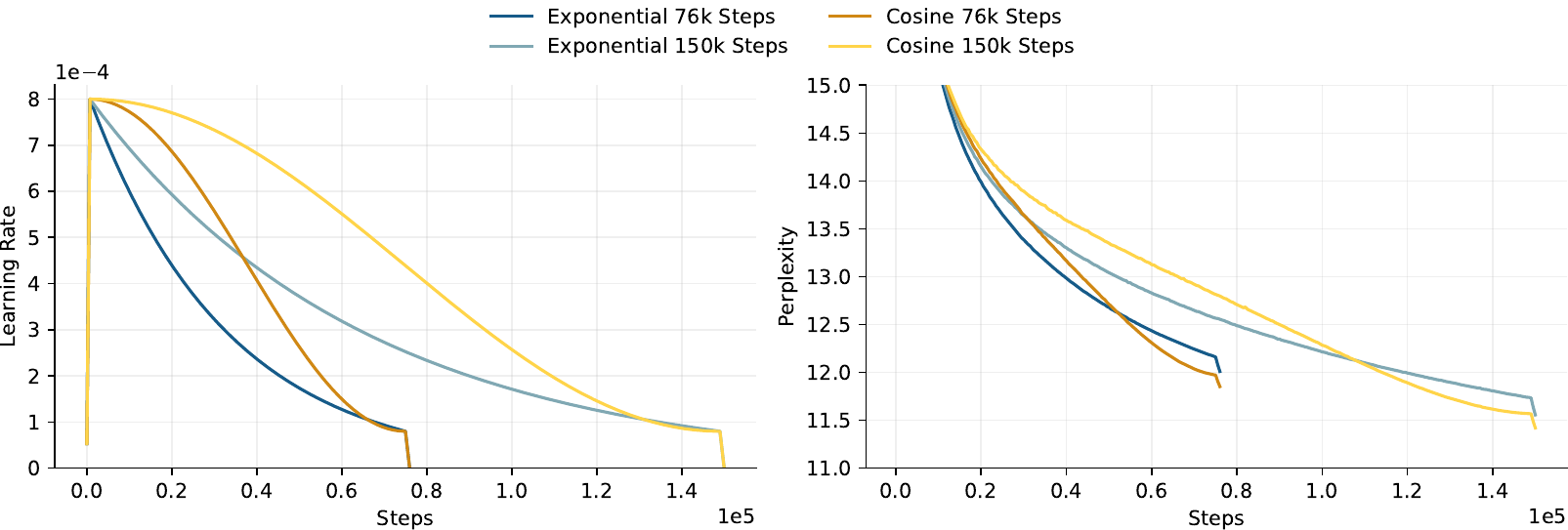}
    \vspace{-0.8em}
    \caption{\label{fig:lr_scheduler}
    Effect of Learning Rate Scheduler. The tested learning rate schedules are shown on the left, with the corresponding training perplexities on the right. 
    While the exponential learning rate schedule can be continued trivially, the cosine schedule actually works slightly better given a fixed number of iterations. The learning rate cooldown to zero at the end gives a similar and significant benefit in both cases.}
\end{figure}

\paragraph{Effect of the Learning Rate Scheduler (Fig.~\ref{fig:lr_scheduler}).} In our largest experiments, we choose a linear warmup followed by an exponential decay as a learning rate schedule in order to enable a continued pre-training with more tokens and without an additional warmup. However, smaller-scale experiments in Fig.~\ref{fig:lr_scheduler} show the benefit of a cosine schedule over an exponential one.

\begin{figure}[H]
    \centering
\includegraphics[width=0.5\columnwidth]{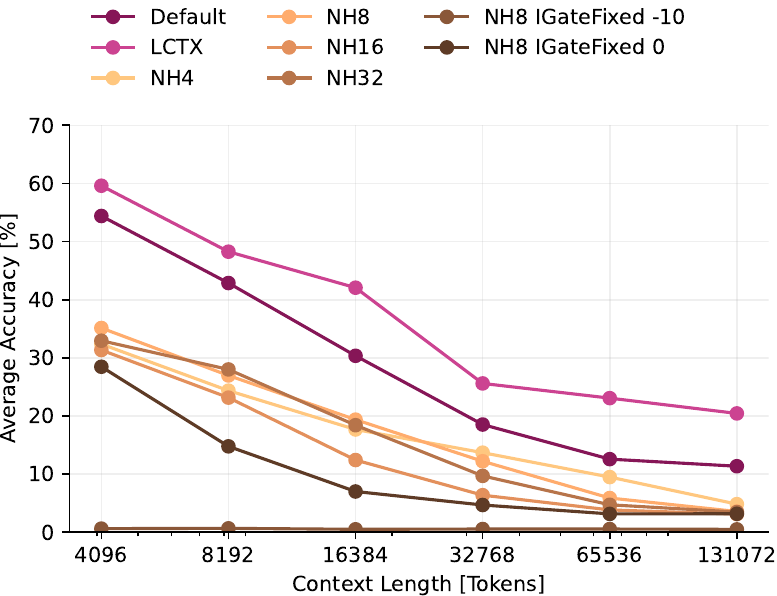}
    \vspace{-0.8em}
    \caption{RULER average accuracies for different number of heads/cell dimensions, and fixed input gate. The ablations are trained on 160B tokens at 8k context.\label{fig:ruler_head_igate_abl}}
\end{figure}

\paragraph{Effect of Memory State Size and Input Gate on Long Context Evaluations (Fig.~\ref{fig:ruler_head_igate_abl}, Tab.~\ref{tab:leaderboardv2_abl}~and~\ref{tab:leaderboardv1_abl}).}
\label{app:longctx_eval_abl}
In order to test the influence of the head numbers (cell dimensions) and input gate on long context abilities, we test the ablation models trained in Sec.~\ref{sec:experiments:ablations} for their performance in the RULER benchmark~\citep{hsieh2024ruler}. 
The results in Fig.~\ref{fig:ruler_head_igate_abl} show that, while the effect of the head number and equivalently the recurrent memory is inconclusive, the models strongly benefit from the learnable, exponential input gate for the long context performance.

Additionally, we evaluate our ablation versions trained for 160B tokens and evaluated on the current and old HuggingFace LLM Leaderboard as in Tab.~\ref{tab:leaderboardv2} and \ref{tab:leaderboardv1}, respectively. Results in Tab.~\ref{tab:leaderboardv2_abl}, \ref{tab:leaderboardv1_abl} show only slight influence of the head dimensions or fixing input gate. Only fixing the input gate to the very small value of its standard bias initialization has a stronger impact on the Leaderboard v1.
\begin{table*}[h]
\caption{Model Performance for different number of heads and non-trainable input gate on the Huggingface Leaderboard v2 tasks. \\ $\uparrow$ indicates larger values are better.}
\label{tab:leaderboardv2_abl}
    \vspace{0.2cm}
    \begin{adjustbox}{width=\textwidth}
    \begin{tabular}{lccccccc}
        \toprule
        \textsc{Model} & \textsc{BBH} $\uparrow$& \textsc{MMLU-Pro} $\uparrow$ & \textsc{Math} $\uparrow$& \textsc{MuSR} $\uparrow$ & \textsc{GPQA} $\uparrow$& \textsc{IFEval} $\uparrow$& \textsc{Average} $\uparrow$\\ 
        \midrule 
\xlstmlarge abl NH4 & 0.306 & 0.114 & 0.004 & 0.363 & 0.253 & 0.160 & 0.200\\ 
\xlstmlarge abl NH8 & 0.304 & 0.115 & 0.002 & 0.363 & 0.248 & 0.173 & 0.201\\ 
\xlstmlarge abl NH16 & 0.317 & 0.119 & 0.002 & 0.390 & 0.258 & 0.161 & 0.208\\ 
\xlstmlarge abl NH32 & 0.327 & 0.120 & 0.001 & 0.379 & 0.256 & 0.171 & 0.209\\ 
\xlstmlarge abl NH8 IGateFixed 0 & 0.303 & 0.117 & 0.004 & 0.381 & 0.229 & 0.149 & 0.197 \\
\xlstmlarge abl NH8 IGateFixed -10 & 0.308 & 0.109 & 0.000 & 0.357 & 0.253 & 0.165 & 0.199\\ 
        \midrule
        \textbf{\xlstmlarge}               & 0.381 & 0.242 & 0.036 & 0.379 & 0.280 & 0.244 & 0.260 \\
        \textbf{\xlstmlarge} LCTX & 0.390 & 0.252 & 0.040 & 0.374 & 0.253 & 0.234 & 0.257 \\
        \bottomrule
    \end{tabular}
    \end{adjustbox}
\end{table*}

\begin{table*}[h]
\caption{Model Performance for different number of heads and non-trainable input gate on the Huggingface Leaderboard v1 tasks. \\
$\uparrow$ indicates larger values are better.}
\label{tab:leaderboardv1_abl}
\vspace{0.2cm}
\begin{adjustbox}{width=\textwidth}
    \begin{tabular}{lcccccccc}
        \toprule
        \textsc{Model} & \textsc{ARC-C} $\uparrow$& \textsc{MMLU} $\uparrow$& \textsc{HellaSwag} $\uparrow$& \textsc{Winogrande}$\uparrow$ & \textsc{TruthfulQA} $\uparrow$& \textsc{OpenBookQA} $\uparrow$& \textsc{PiQA} $\uparrow$& \textsc{Average} $\uparrow$\\ 
\midrule
\xlstmlarge abl NH4 & 0.492 & 0.296 & 0.665 & 0.672 & 0.282 & 0.405 & 0.798 & 0.516\\ 
\xlstmlarge abl NH8 & 0.487 & 0.292 & 0.669 & 0.680 & 0.302 & 0.426 & 0.791 & 0.521\\ 
\xlstmlarge abl NH16 & 0.505 & 0.351 & 0.668 & 0.701 & 0.294 & 0.409 & 0.796 & 0.532\\ 
\xlstmlarge abl NH32 & 0.500 & 0.378 & 0.666 & 0.676 & 0.325 & 0.411 & 0.799 & 0.536\\ 
\xlstmlarge abl NH8 IGateFixed 0 & 0.464 & 0.292 & 0.658 & 0.672 & 0.280 & 0.415 & 0.788 & 0.510\\ 
\xlstmlarge abl NH8 IGateFixed -10 & 0.241 & 0.250 & 0.340 & 0.519 & 0.286 & 0.226 & 0.681 & 0.363\\ 
\midrule 
\textbf{\xlstmlarge}  & 0.574 & 0.578 & 0.714 & 0.738 & 0.419 & 0.448 & 0.819 & 0.613\\ 
\textbf{\xlstmlarge} LCTX & 0.516 & 0.588 & 0.715 & 0.740 & 0.374 & 0.429 & 0.819 & 0.597\\ 

        \bottomrule
    \end{tabular}
    \end{adjustbox}
\end{table*}

\section{FLOP Counting}\label{app:flop_counting}

    We count the number of FLOPs in a forward pass of the mLSTM. We use a factor of 2 to describe the multiply accumulate cost.

    We use factors denoted as F\_X to describe the number of FLOPs for operation X (e.g. F\_exp for the exponential function).
    By default we set all of these factors to 1.

    \subsection{FLOPs for the mLSTM Operation}

    \begin{itemize}
        \item Inter-chunk recurrent:
              \begin{itemize}
                  \item \textbf{Chunkwise gates:}  num\_heads $\times$ num\_chunks \\
                        $\times$ ( 0.5$\times$chunk\_size $\times$ (chunk\_size + 1) + 2$\times$chunk\_size )
                  \item \textbf{Gates \& max state:} num\_heads $\times$ num\_chunks \\
                        $\times$ ( 3 + F\_max + F\_exp + chunk\_size $\times$ (3 + 2 $\times$ F\_exp))
                  \item \textbf{Numerator:} num\_heads $\times$ num\_chunks\\
                        $\times$ (2$\times$d\_qk $\times$ d\_v + 4$\times$chunk\_size $\times$ d\_qk $\times$ d\_v + 3$\times$chunk\_size $\times$ d\_qk)
                  \item \textbf{Denominator:} num\_heads $\times$ num\_chunks $\times$ ( d\_qk + 4$\times$chunk\_size $\times$ d\_qk )
              \end{itemize}
        \item Intra-chunk parallel:
              \begin{itemize}
                  \item \textbf{Gate matrix:} num\_heads $\times$ num\_chunks \\
                  $\times$ ( 0.5 $\times$ chunk\_size $\times$ (chunk\_size + 1) \\
                  + chunk\_size $\times$ chunk\_size $\times$ (3 + F\_mask + F\_max + F\_exp) \\
                  + chunk\_size $\times$ (1 + F\_max) )
                  \item \textbf{Gated Attn logits:}  num\_heads $\times$ num\_chunks \\
                  $\times$ 2$\times$chunk\_size $\times$ chunk\_size $\times$ ( 1 + d\_qk )
                  \item \textbf{Numerator:}  num\_heads $\times$ num\_chunks \\
                  $\times$ 2$\times$chunk\_size $\times$ chunk\_size $\times$ d\_v
                  \item \textbf{Denominator:}  num\_heads $\times$ num\_chunks $\times$ 2 $\times$ chunk\_size $\times$ chunk\_size
                  \item \textbf{Output combination:}  num\_heads $\times$ num\_chunks \\
                  $\times$ ( chunk\_size $\times$ ( 1 + F\_max ) \\ + chunk\_size $\times$ ( 2 + F\_abs + F\_exp + F\_max + 2$\times$d\_v ) )
              \end{itemize}
    \end{itemize}

    \subsection{FLOPs for the mLSTM in a Transformer Backbone}
    \label{secapp:flops_mlstm}
    For computing the number of FLOPs we follow the procedure from~\citet{Hoffmann2022Chinchilla}.
    We include the FLOPs contributed by the embedding matrices. We do not include RMS- or Layer-Norm and skip connection FLOPs
    We assume that the backward pass has 2 times the number of FLOPs of the forward pass.
    For the forward pass, the number of FLOPs of the mLSTM for a single sequence can be approximated by:

    \begin{itemize}
        \item Embeddings
              \begin{itemize}
                  \item 2 $\times$ seq\_len $\times$ vocab\_size $\times$ d\_model
              \end{itemize}
        \item mLSTM (single layer)
              \begin{itemize}
                  \item \textbf{Query, key, value, input and forget gate projections:} \\
                        2 $\times$ seq\_len $\times$ d\_model $\times$ num\_heads $\times$ (2 $\times$ d\_qk + d\_v + 2)
                  \item \textbf{Output gate and projection:} \\
                        4 $\times$ seq\_len $\times$ d\_model $\times$ num\_heads $\times$ d\_v \\
                        + seq\_len $\times$ num\_heads $\times$ d\_v $\times$ F\_sig
                  \item \textbf{mLSTM cell:} See above.
              \end{itemize}
        \item Gated Feedforward (single layer)
              \begin{itemize}
                  \item 6 $\times$ seq\_len $\times$ d\_model $\times$ d\_model $\times$ proj\_factor\_ff \\
                        + 2 $\times$ seq\_len $\times$ d\_model $\times$ F\_swish
              \end{itemize}
        \item Final Logits
              \begin{itemize}
                  \item 2 $\times$ seq\_len $\times$ d\_model $\times$ vocab\_size
              \end{itemize}
        \item \textbf{Total forward pass FLOPs:} \\
              embeddings + num\_layers $\times$ (mLSTM + feedforward) + final\_logits
    \end{itemize}

    \subsection{FLOPs for the Transformer with Self-Attention}
    We use the FLOP computations from~\citet{Hoffmann2022Chinchilla}, with the difference that we use gated feedforward blocks.

    \begin{itemize}
        \item Embeddings
              \begin{itemize}
                  \item 2 $\times$ seq\_len $\times$ vocab\_size $\times$ d\_model
              \end{itemize}
        \item Attention (single layer)
              \begin{itemize}
                  \item \textbf{Key, query and value projections:} \\
                        2 $\times$ seq\_len $\times$ d\_model $\times$ num\_heads $\times$ (2 $\times$ d\_qk + d\_v)
                  \item \textbf{Key @ query logits:} 2 $\times$ seq\_len $\times$ seq\_len $\times$ (d\_qk $\times$ num\_heads)
                  \item \textbf{Softmax:} 3 $\times$ seq\_len $\times$ seq\_len $\times$ num\_heads
                  \item \textbf{Softmax @ query reductions:} 2 $\times$ seq\_len $\times$ seq\_len $\times$ (num\_heads $\times$ d\_qk)
                  \item \textbf{Final linear:} 2 $\times$ seq\_len $\times$ d\_model $\times$ (num\_heads $\times$ d\_v)
              \end{itemize}
        \item Gated Feedforward (single layer)
              \begin{itemize}
                  \item 6 $\times$ seq\_len $\times$ d\_model $\times$ d\_model $\times$ proj\_factor\_ff \\
                        + 2 $\times$ seq\_len $\times$ d\_model $\times$ F\_swish
              \end{itemize}
        \item Final Logits
              \begin{itemize}
                  \item 2 $\times$ seq\_len $\times$ d\_model $\times$ vocab\_size
              \end{itemize}
        \item \textbf{Total forward pass FLOPs:} \\
              embeddings + num\_layers $\times$ (attention + feedforward) + final\_logits
    \end{itemize}

    \section{Parameter Counting}
    In this section we count the number of paramters in the mLSTM and compare it to the number of parameters
    in a Transformer with self-attention.
    We assume that the model does not use weight tying and omits biases.
    \subsection{Parameter Counting for the mLSTM}

    \begin{itemize}
        \item Embeddings
              \begin{itemize}
                  \item vocab\_size $\times$ d\_model
              \end{itemize}
        \item mLSTM (single layer)
              \begin{itemize}
                  \item \textbf{qkv:} d\_model $\times$ num\_heads $\times$ (2 $\times$ d\_qk + d\_v)
                  \item \textbf{Input and forget gate:} 2 $\times$ d\_model $\times$ num\_heads + 2 $\times$ num\_heads
                  \item \textbf{Output gate:} d\_model $\times$ d\_model
                  \item \textbf{Output projection:} d\_model $\times$ d\_model
                  \item \textbf{Norm:} d\_model
              \end{itemize}
        \item Gated Feedforward (single layer)
              \begin{itemize}
                  \item 3 $\times$ d\_model $\times$ d\_model $\times$ proj\_factor\_ff
              \end{itemize}
        \item Norm (single layer)
              \begin{itemize}
                  \item d\_model
              \end{itemize}
        \item Final Logits:
              \begin{itemize}
                  \item d\_model $\times$ vocab\_size
              \end{itemize}
        \item \textbf{Total number of parameters:} \\
              embeddings + num\_layers $\times$ (mLSTM + feedforward + 2 $\times$ norm) + norm + final\_logits
    \end{itemize}

    \subsection{Parameter Counting for the Transformer with Self-Attention}

    \begin{itemize}
        \item Embeddings
              \begin{itemize}
                  \item vocab\_size $\times$ d\_model
              \end{itemize}
        \item Attention (single layer)
              \begin{itemize}
                  \item \textbf{qkv:} d\_model $\times$ num\_heads $\times$ (2 $\times$ d\_qk + d\_v)
                  \item \textbf{Output projection:} d\_model $\times$ d\_model
              \end{itemize}
        \item Gated Feedforward (single layer)
              \begin{itemize}
                  \item 3 $\times$ d\_model $\times$ d\_model $\times$ proj\_factor\_ff
              \end{itemize}
        \item Norm (single layer)
              \begin{itemize}
                  \item d\_model
              \end{itemize}
        \item Final Logits:
              \begin{itemize}
                  \item d\_model $\times$ vocab\_size
              \end{itemize}
        \item \textbf{Total number of parameters:} \\
              embeddings + num\_layers $\times$ (attention + feedforward + 2 $\times$ norm) + norm + final\_logits
    \end{itemize}

\end{document}